\documentclass[12pt]{article}
\usepackage[top=1in, bottom=1in, left=1in, right=1in]{geometry}

\usepackage{cite}
\usepackage{amsmath,amssymb,amsfonts}
\usepackage{algorithmic}
\usepackage{graphicx}
\usepackage{textcomp}
\usepackage{xcolor}
\usepackage{adjustbox}
\usepackage{booktabs}
\usepackage{multirow}
\usepackage{subcaption}
\usepackage{pgfplots}
\usepackage[absolute]{textpos}
\usepackage{float}
\usepackage{fancyhdr}

\begin{document}

\vspace{-1.5em}
\begin{flushright}
\end{flushright}

\begin{center}
    \Large \textbf{Intelligent Incident Hypertension Prediction in
Obstructive Sleep Apnea}
    
    \vspace{1em}
    \normalsize
    Omid Halimi Milani\textsuperscript{1},
    Ahmet Enis Cetin\textsuperscript{1}, Bharati Prasad\textsuperscript{2} \\
    
    \vspace{0.5em}
    \textsuperscript{1}Department of Electrical and Computer Engineering, University of Illinois Chicago, IL, USA \\
    \textsuperscript{2}Department of Medicine, Division of Pulmonary, Critical Care, \\ Sleep and Allergy, University of Illinois at Chicago, Chicago, IL, USA \\

\end{center}

\begin{abstract}
Obstructive sleep apnea (OSA) is a significant risk factor for hypertension, primarily due to intermittent hypoxia and sleep fragmentation. Predicting whether individuals with OSA will develop hypertension within five years remains a complex challenge. This study introduces a novel deep learning approach that integrates Discrete Cosine Transform (DCT)-based transfer learning to enhance prediction accuracy. We are the first to incorporate all polysomnography signals together for hypertension prediction, leveraging their collective information to improve model performance. Features were extracted from these signals and transformed into a 2D representation to utilize pre-trained 2D neural networks such as MobileNet, EfficientNet, and ResNet variants. To further improve feature learning, we introduced a DCT layer, which transforms input features into a frequency-based representation, preserving essential spectral information, decorrelating features, and enhancing robustness to noise. This frequency-domain approach, coupled with transfer learning, is especially beneficial for limited medical datasets, as it leverages rich representations from pre-trained networks to improve generalization. By strategically placing the DCT layer at deeper truncation depths within EfficientNet, our model achieved a best area under the curve (AUC) of 72.88\%, demonstrating the effectiveness of frequency-domain feature extraction and transfer learning in predicting hypertension risk in OSA patients over a five-year period.

\end{abstract}
%
%
\section{Introduction}
Obstructive sleep apnea (OSA) is a condition characterized by intermittent hypoxia and sleep fragmentation, which propagates hypertension via mechanisms such as sympathetic activation and inflammation. OSA also raises the likelihood of developing hypertension during nighttime \cite{peppard2000prospective, dudenbostel2012resistant}.


Notably, specific patterns of apnea, such as Rapid Eye Movement (REM) sleep-related OSA, could contribute to OSA-associated hypertension \cite{mokhlesi2014obstructive}. Additional sleep-related factors, like reduced slow-wave sleep (SWS) and short sleep duration, are associated with hypertension, independent of OSA \cite{javaheri2018slow, wang2021association}.

Ren and colleagues conducted a study examining the association between sleep duration, obstructive sleep apnea (OSA), and hypertension in a group of 7,107 OSA patients and 1,118 primary snorers. The findings from polysomnography indicated that individuals who slept for 5 to 6 hours had a 45\% risk of developing hypertension, while those who slept fewer than 5 hours had an 80\% increased risk, independent of other influencing factors \cite{ren2018objective}.

Treating obstructive sleep apnea (OSA) has been shown to lower the risk of developing hypertension \cite{marin2012association}. However, accurately predicting the onset of hypertension in individuals with OSA remains a challenge due to the complex pathogenesis of hypertension.

This study leveraged deep Machine Learning (ML) techniques to develop a predictive model for incident hypertension up to five years after OSA diagnosis by polysomnography. The novel ML-based method utilized polysomnography signal conversion and a composite deep CNN model to transform the polysomnography signals into image datasets to predict incident hypertension. During the training process, additional static information for each patient was incorporated, including age, sex, race, body mass index, baseline systolic, and diastolic blood pressure. Our study aimed to accelerate precision medicine in managing hypertension and preventive healthcare by integrating polysomnography and clinical data analytics.

We hypothesized that a comprehensive approach involving the simultaneous input of time-series physiological signals measuring sleep (EEG), ventilatory impairment and hypoxia, and cardiac autonomic dysregulation (electrocardiogram and photoplethysmography-derived heart rate variability and pulse transit time) could preserve the temporal correlations between multiple physiological perturbations in OSA and provide a robust prediction of incident hypertension \cite{zheng2024semi}. These signals were included in polysomnography, a widely available diagnostic test for OSA. Thus, we extracted multiple features from the polysomnography signals in the SHHS participants with moderate to severe OSA \cite{quan1997sleep, zhang2018national}. 


Before feature extraction, rigorous preprocessing steps were employed, involving removing artifacts and applying a bandpass filter to eliminate any remaining noise.

The proposed methodology underwent a rigorous evaluation through a 10-fold cross-validation approach to examine the model’s generalizability. The results were subsequently summarized, comparing models and methods to cutting-edge approaches.

\noindent
Our main contributions are summarized as follows:
\begin{itemize}

    \item We develop a DCT-based convolution framework that replaces the complex-valued DFT with a real-valued, orthogonal DCT, thereby simplifying convolution operations and preserving crucial frequency information in polysomnography signals.
    \item We introduce threshold-based nonlinearities (soft and hard thresholding) in the DCT domain, preventing the loss of important negative-frequency coefficients that standard ReLU would discard.
    \item We embed the DCT layer within a truncated EfficientNet architecture, using time-windowed feature extraction and transfer learning to efficiently process multi-channel polysomnography data.
    \item We demonstrate that this end-to-end model accurately predicts long-term hypertension risk, providing a robust framework for precision healthcare in OSA.

\end{itemize}

\section{Related Works}
In recent years, there has been a growing recognition of the limitations associated with conventional manual sleep stage scoring, which simplifies the analysis of electroencephalogram (EEG) temporospectral and frequency domains. This scoring method is inherently subjective and can lead to variations between different scorers due to the application of visual-based rules \cite{schulz2008rethinking}. Researchers used power spectral density (PSD) analysis of EEG signals, which examines sleep EEG microarchitecture. This approach enables the decomposition of EEG brain waves across various power frequency bands, ranging from slow wave activity (delta EEG power, 1–4 Hz) to fast-frequency activity (beta EEG power, 18–30 Hz), achieved through fast Fourier transform algorithms. At a microarchitecture level, slow wave sleep (SWS) is characterized by high delta power, indicative of deep sleep. Quantitative EEG analysis may yield more sensitive biomarkers for adverse health outcomes in OSA compared to traditional sleep scoring methods \cite{ appleton2019quantitative,d2017quantitative, lechat2022novel, parker2022association}.

Berger et al. recently reported that low delta power in non-REM sleep is independently associated with the risk of developing hypertension, confirming the previously noted relationship of SWS with incident hypertension \cite{berger2022association}. Additionally, a recent study proposed focusing exclusively on oxygen saturation (SpO\textsubscript{2}) features for predicting incident hypertension in OSA patients using time and frequency domain analysis, achieving 84.4\% accuracy, 77.0\% sensitivity, 91.5\% specificity, and an AUC of 84.3\% with a random forest model \cite{you2023risk}.

Ruitong et al. adopted a pulmonary physiology-based approach to predicting the onset of hypertension by including pulmonary function measurements and polysomnography-derived indices using a penalized regression and Elastic Net model \cite{li2021composite}. A recent study introduced cSP (sleep and pulmonary) phenotypes, which combines spirometry and overnight polysomnography measures to predict hypertension occurrence in the Sleep Heart Health Study (SHHS) \cite{appleton2019quantitative}. The SHHS dataset encompasses a variety of physiological signals, including sleep EEG, electrocardiogram (ECG), electromyogram (EMG), ventilatory effort, nasal airflow, photoplethysmography-derived oximetry, snoring, and body position. Employing rigorous signal processing techniques, such as filtering, segmentation, and feature extraction, our analysis aimed to unravel the intricate patterns embedded within these signals \cite{quan1997sleep, zhang2018national}.

\section{Methodology}
We propose a feature-based deep learning method for predicting incident hypertension in patients with obstructive sleep apnea (OSA). Convolution operations are typically performed in the time (or spatial) domain, but they can also be implemented more efficiently in the frequency domain via element-wise multiplication. Let \(y = w * x\) be the convolution of a filter \(w\) and signal \(x\). By applying the Discrete Fourier Transform (DFT), we obtain:
\begin{equation}
\label{eq:dft_conv}
Y[k] = W[k] \cdot X[k],
\end{equation}
where \(Y\), \(W\), and \(X\) are the DFTs of \(y\), \(w\), and \(x\), respectively. However, the DFT produces complex-valued coefficients, complicating the use of conventional non-linearities such as ReLU. To avoid this, we use the Discrete Cosine Transform (DCT), a purely real and orthogonal transform, which preserves signal energy in the frequency domain while reducing redundancy.

\begin{figure}[htbp]
    \centering
    \includegraphics[width=0.5\linewidth]{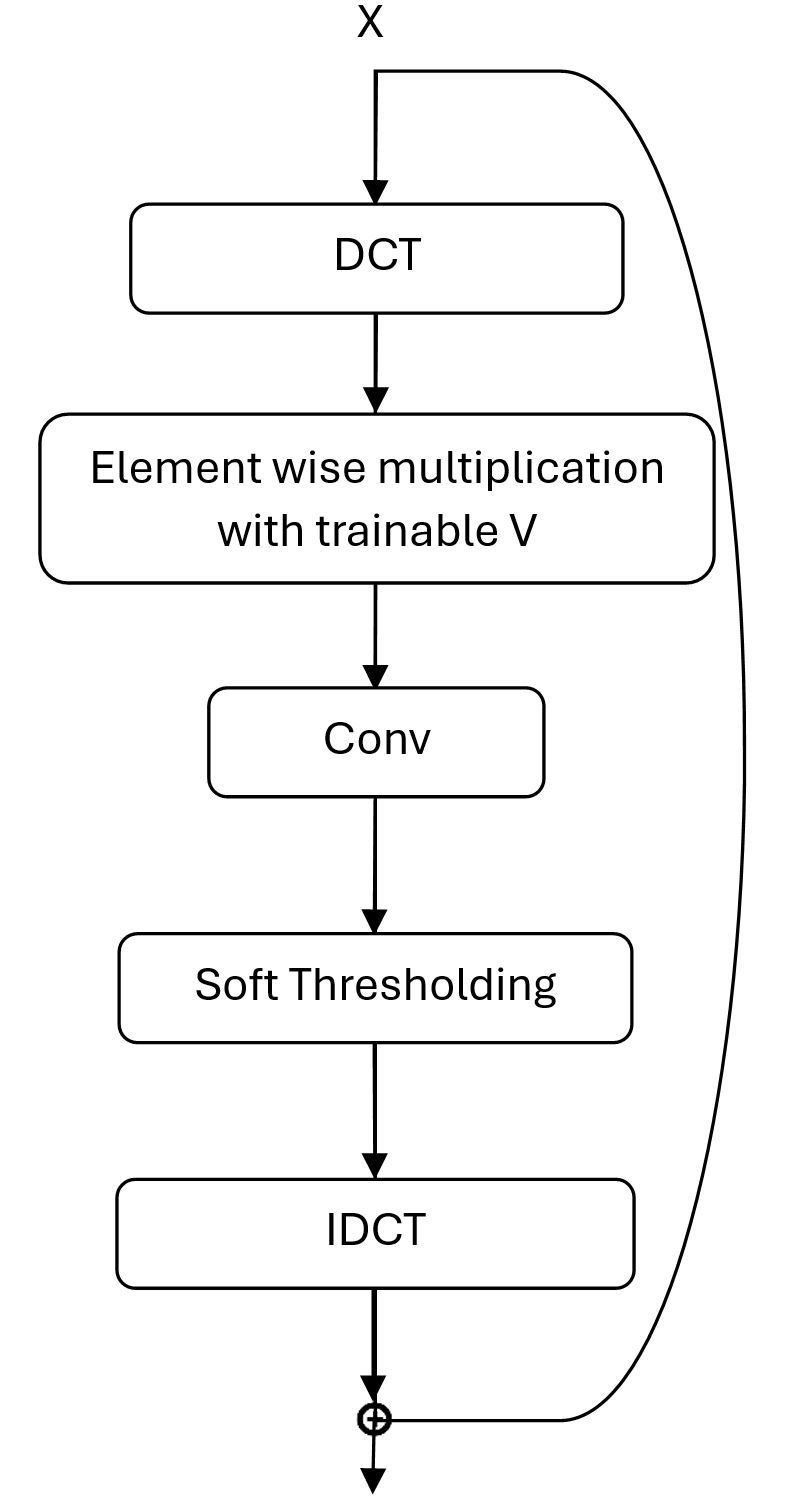} 
    \caption{The structure of the DCT block.}
    \label{fig:dct_block}
\end{figure}

\vspace{1em}
\noindent
A key feature of our approach is the replacement of traditional convolutions with a DCT-based convolution block (illustrated in Fig.~\ref{fig:dct_block}). Rather than applying ReLU activations in the DCT domain—which would discard negative-frequency information—we incorporate a soft thresholding operator. Originally introduced in image denoising, soft thresholding effectively prunes smaller (potentially noisy) DCT coefficients while retaining both positive and negative high-amplitude components. In our PyTorch implementation, it is defined as:
\begin{equation}
\label{eq:softthresh}
\mathrm{SoftThreshold}(x,\tau) = \mathrm{sign}(x)\,\max\bigl(\lvert x\rvert - \tau, 0\bigr).
\end{equation}

\noindent
By applying soft thresholding in place of ReLU, we preserve crucial frequency details that might otherwise be lost, thereby improving the expressive capacity of the model.

\vspace{1em}
\noindent
Another important characteristic of the DCT layer is its orthogonality. Because each transform basis (cosine wave) is orthogonal to the others, the resulting representation can decorrelate input features, often leading to improved generalization and numerical stability compared to complex-valued transforms.

\vspace{1em}
\noindent
To reduce computational overhead, we do not operate on raw polysomnography signals directly. Instead, each signal (EEG, ECG, respiratory, etc.) is first preprocessed (e.g., filtered at relevant passbands) and then segmented into time windows. From each window, we extract a set of representative features, including:
\begin{itemize}
    \item Counts and durations of respiratory events (apneas, hypopneas) and arousals,
    \item Statistical descriptors (mean, standard deviation, skewness, kurtosis),
    \item Heart rate variability (HRV) metrics derived from ECG signals.
\end{itemize}

\vspace{1em}
\noindent
\textbf{HRV Feature Computation.}
After detecting R-peaks within each window, we obtain the sequence of R-R intervals, \(\{RR_i\}_{i=1}^{N}\). Let
\(\overline{RR}\) be the mean R-R interval:
\begin{equation}
\label{eq:rr_mean}
\overline{RR} = \frac{1}{N}\sum_{i=1}^{N} RR_i.
\end{equation}

\noindent
We then compute the following standard HRV metrics:

\begin{equation}
\label{eq:sdnn}
\mathrm{SDNN} = \sqrt{\frac{1}{N} \sum_{i=1}^{N} \bigl(RR_i - \overline{RR}\bigr)^{2}},
\end{equation}

\begin{equation}
\label{eq:rmssd}
\mathrm{RMSSD} = \sqrt{\frac{1}{N-1} \sum_{i=1}^{N-1} \bigl(RR_{i+1} - RR_{i}\bigr)^{2}},
\end{equation}

\begin{equation}
\label{eq:pnn50}
\mathrm{pNN50} = \frac{\mathrm{NN50}}{N-1} \times 100\%,
\end{equation}

\begin{equation}
\label{eq:nn50}
\mathrm{NN50} = \sum_{i=1}^{N-1} \mathbf{1}\bigl(\lvert RR_{i+1} - RR_{i}\rvert > 50\text{ ms}\bigr).
\end{equation}

\noindent
Here, \(N\) is the number of valid R-R intervals within the window. \(\mathrm{SDNN}\) measures overall variability, \(\mathrm{RMSSD}\) captures short-term fluctuations, and \(\mathrm{pNN50}\) reflects the percentage of adjacent R-R intervals that differ by more than 50\,ms.

\vspace{1em}
\noindent
We concatenate all window-level features into a \(2D\) array (\(\text{features}\times\text{windows}\)), treating it as a pseudo-image \cite{milani20230537}. This format permits us to use 2D CNN architectures while leveraging the DCT layer for frequency-domain transformations.


\vspace{1em}
\noindent
By replacing DFT-based convolutions with a real-valued DCT transform and using soft thresholding as a nonlinearity, we retain essential negative-frequency information that might otherwise be discarded by ReLU. Our proposed method thus combines the efficiency of feature-level input, the power of a truncated CNN backbone, and the spectral advantages of an orthogonal DCT layer, offering an effective framework for predicting incident hypertension in OSA patients.

\section{Results}

To evaluate the performance of our \emph{feature-based} approach, we employed 10-fold cross-validation with various time window lengths ranging from 9 to 60 minutes. We present here the results for 60-minute and 10-minute intervals, as well as a brief investigation into optimized shorter windows around 10 minutes. We then discuss how adding a Discrete Cosine Transform (DCT) block at different depths within a truncated EfficientNet affects performance.

\subsection{60-Minute Interval Results}
Table~\ref{tab:60min} summarizes the accuracy and AUC for the 60-minute interval. Among the feature-based models, Feature\_EffNet-B0 achieved the highest accuracy (\(68.66\%\)) and AUC (\(68.66\%\)). 
Other models, such as Feature\_ResNet-18 and Feature\_MobileNet-v2, provided lower sensitivity but stronger specificity, effectively minimizing false positives while missing more true positives.


\begin{table}[htbp]
    \centering
    \caption{Performance of Models at the 60-Minute Interval}
    \label{tab:60min}
    \begin{tabular}{lcc}
        \hline
        \textbf{Model} & \textbf{Accuracy (\%)} & \textbf{AUC (\%)} \\
        \hline
        \textbf{Feature\_EffNet-B0}  & \textbf{68.66} & \textbf{68.66} \\
        Feature\_ResNet-10  & 62.69 & 59.70 \\
        Feature\_ResNet-18  & 61.19 & 55.22 \\
        Feature\_MobileNet-v2 & 65.67 & 61.19 \\
        \hline
    \end{tabular}
\end{table}

\subsection{10-Minute Interval Results}
A shorter 10-minute window offers more granular insight into sleep data. As shown in Table~\ref{tab:10min}, Feature\_EffNet-B0 achieved the highest accuracy (\(68.66\%\)) and AUC (\(71.64\%\)). 

Incorporating static features similarly provided small boosts in certain metrics. Overall, these results suggest that a 10-minute window balances capturing sufficient polysomnography patterns with reduced data overhead.

\begin{table}[htbp]
    \centering
    \caption{Performance of Models at the 10-Minute Interval}
    \label{tab:10min}
    \begin{tabular}{lcc}
        \hline
        \textbf{Model} & \textbf{Accuracy (\%)} & \textbf{AUC (\%)} \\
        \hline
        Feature\_EffNet-B0  & \textbf{68.66} & \textbf{71.64} \\
        Feature\_ResNet-10  & 62.69 & 60.70 \\
        Feature\_ResNet-18  & 61.19 & 57.31 \\
        Feature\_MobileNet-v2 & 65.67 & 63.43 \\
        \hline
    \end{tabular}
\end{table}



\subsection{DCT Block Placement at Different Truncation Depths}
We evaluated how inserting the DCT2D layer at different truncation depths in EfficientNet-B0 affects predictive performance. Specifically, we integrated the DCT block after the third, fourth, fifth, or sixth building block in the network. Table~\ref{tab:dct_depths} presents the best observed accuracy and AUC for each placement.

\begin{table}[htbp]
    \centering
    \caption{Best Accuracy and AUC by Inserting the DCT2D Block at Different EfficientNet-B0 Depths}
    \label{tab:dct_depths}
    \begin{tabular}{lcc}
        \hline
        \textbf{Model} & \textbf{Accuracy (\%)} & \textbf{AUC (\%)} \\
        \hline
        DCT@3 (after 3rd block) & 68.66 & 70.81 \\
        DCT@4 (after 4th block) & 69.66 & 70.73 \\
        DCT@5 (after 5th block) & 68.12 & 72.79 \\
        DCT@6 (after 6th block) & 69.86 & 72.88 \\
        \hline
    \end{tabular}
\end{table}

As shown in Table~\ref{tab:dct_depths}, placing the DCT layer at deeper levels (DCT@5 or DCT@6) yields higher AUC values (72.29\% and 72.88\%, respectively), with DCT@6 also achieving the top accuracy (69.86\%). This suggests that mid- to late-stage feature maps in EfficientNet-B0 may benefit more from the frequency-domain transform, likely due to increasingly abstract representations of the polysomnography data at deeper layers. Ultimately, the choice between prioritizing accuracy or maximizing AUC may depend on clinical considerations, such as avoiding false negatives versus improving overall predictive discrimination.

\subsection{Comparison with State-of-the-Art Methods}
Table~\ref{tab:sota} compares our best-performing DCT-based approach to established methods, including cSPPSG and AHI \cite{li2021composite}. Incorporating the DCT layer at depth 5 or 6 outperforms these baselines in terms of AUC, underscoring the advantage of combining frequency-domain transformations with threshold-based nonlinearities for improved hypertension-risk prediction in OSA.

\begin{table}[htbp]
    \centering
    \caption{Comparison with State-of-the-Art Models}
    \label{tab:sota}
    \begin{tabular}{lcc}
        \hline
        \textbf{Model} & \textbf{Accuracy (\%)} & \textbf{AUC (\%)} \\
        \hline
        cSPPSG \cite{li2021composite} & - & 71 \\
        AHI \cite{li2021composite} & - & 67 \\
        \textbf{Ours}  & 69.86 & \textbf{72.88} \\
        \hline
    \end{tabular}
\end{table}

In conclusion, integrating the DCT layer at deeper truncation depths within EfficientNet-B0 yields meaningful gains in AUC while maintaining high accuracy. This finding highlights the synergy between advanced CNN architectures and frequency-domain transformations for predicting long-term hypertension in OSA patients.

\section{Conclusion}

This study introduced a DCT-enhanced deep learning framework to predict incident hypertension in obstructive sleep apnea (OSA) patients using polysomnography data. By embedding threshold-based nonlinearities in the DCT domain within a truncated EfficientNet backbone, we preserved essential negative-frequency information and effectively leveraged multi-signal features. Our best-performing configuration achieved 69.86\% accuracy and a 72.88\% AUC, surpassing existing models such as cSPPSG and AHI, which achieved 71\% and 67\% AUC respectively. These results underscore the potential of combining frequency-domain transformations, transfer learning, and comprehensive feature extraction to address the complex mechanisms linking OSA and hypertension.

Despite these promising findings, the limited dataset size may impact generalizability. Future work will focus on expanding data availability, and further improving the model architecture. By addressing these areas, we aim to enhance predictive performance and support earlier, more precise interventions for OSA patients at risk of developing hypertension.

\vspace{1.5cm} 


\end{document}